\pdfoutput=1
\documentclass[9pt,conference]{IEEEtran}
\IEEEoverridecommandlockouts
\AtBeginDocument{%
  \providecommand\BibTeX{{%
    \normalfont B\kern-0.5em{\scshape i\kern-0.25em b}\kern-0.8em\TeX}}}

\usepackage{threeparttable}
\usepackage{titlesec}
\titlespacing{\section}{0pt}{-2pt}{-2pt}
\titlespacing{\subsection}{0pt}{-2pt}{-2pt}
\titlespacing{\subsubsection}{0pt}{-2pt}{-2pt}

\usepackage{pifont}
\usepackage{multirow}

\usepackage{flushend}
\usepackage{threeparttable}
\usepackage{cite}
\usepackage{amsmath,amssymb,amsfonts}
\usepackage{algorithmic}
\usepackage{graphicx}
\usepackage{textcomp}
\usepackage{xcolor}
\usepackage{enumitem}
\usepackage[utf8]{inputenc} 
\usepackage[T1]{fontenc}    
\usepackage{hyperref}       
\hypersetup{
    colorlinks=true,
    linkcolor=blue,
    citecolor=blue,
    filecolor=magenta,      
    urlcolor=cyan,
    pdftitle={Overleaf Example},
    pdfpagemode=FullScreen,
    }
    
\usepackage{url}            
\usepackage{booktabs}       
\usepackage{amsfonts}       
\usepackage{nicefrac}       
\usepackage{microtype}      
\usepackage{xcolor}         
\usepackage{graphicx}
\usepackage{booktabs} 
\usepackage{amsmath}
\usepackage{makecell}

\usepackage{url}  
\usepackage{multirow}
\usepackage{amsmath}
\usepackage{amsthm}
\usepackage{pifont}

\usepackage{flushend}

\usepackage{balance}

\begin{document}

\title{A Physics-guided Generative AI Toolkit for \\ Geophysical Monitoring}

\author{
\IEEEauthorblockN{
Junhuan Yang$^{1}$ \quad
Hanchen Wang$^{2}$ \quad
Yi Sheng$^{1}$ \quad
Youzuo Lin$^{2, 3}$ \quad
Lei Yang$^{1}$ \quad
}
\IEEEauthorblockA{
\normalsize
$^{1}$ George Mason University\\
$^{2}$ Los Alamos National Laboratory\\
$^{3}$ University of North Carolina at Chapel Hill
}
}

\maketitle

\begin{abstract}
\noindent Full-waveform inversion (FWI) plays a vital role in geoscience to explore the subsurface. It utilizes the seismic wave to image the subsurface velocity map. As the machine learning (ML) technique evolves, the data-driven approaches using ML for FWI tasks have emerged, offering enhanced accuracy and reduced computational cost compared to traditional physics-based methods. However, a common challenge in geoscience, the unprivileged data, severely limits ML effectiveness. The issue becomes even worse during model pruning, a step essential in geoscience due to environmental complexities. To tackle this, we introduce the EdGeo toolkit, which employs a diffusion-based model guided by physics principles to generate high-fidelity velocity maps. The toolkit uses the acoustic wave equation to generate corresponding seismic waveform data, facilitating the fine-tuning of pruned ML models. Our results demonstrate significant improvements  in SSIM scores and reduction in both  MAE and MSE across various pruning ratios. Notably, the ML model fine-tuned using data generated by EdGeo yields superior quality of velocity maps, especially in representing unprivileged features, outperforming other existing methods.

\end{abstract}

\vspace{3pt}
\section{Introduction}
\begin{figure}[t]
\begin{center}
\includegraphics[width=\columnwidth]{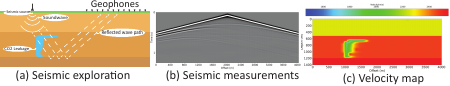}
\caption{An example of (a) Seismic exploration (b) Seismic data and its corresponding (c) Velocity map.}
\label{fig:example}
\end{center}
\end{figure}

Seismic Full-Waveform Inversion (FWI) stands as a cornerstone in the realm of geophysics, employing seismic data processing to unravel intricate details of the subsurface. Its significance lies in its ability to provide high-resolution images, aiding in the characterization of 
potential subsurface hazards \cite{tran2013sinkhole}. A specific application recently raised 
is used to monitor the CO$_2$. Geologic carbon sequestration, a strategy aimed at combating climate change, involves injecting and storing CO2 into deep reservoirs \cite{alumbaugh2023kimberlina}.
The urgency of this endeavor is highlighted by the recent initiation of the Science-informed Machine Learning for Accelerating Real-Time Decisions in Subsurface Applications (SMART) by the US Department of Energy (DOE) \cite{alumbaugh2023kimberlina}. 
This underscores the need for real-time monitoring and decision-making in such applications.
Fig.\ref{fig:example} shows an example of utilizing the seismic data from all seismic signal receivers to produce the underground structure velocity map (Fig.\ref{fig:example} (c)).

Generally, there are 2 ways to implement FWI, physics-driven and data-driven. The physics-driven method produces the velocity map through the physics theories with costly computation and surfers from unsatisfied performance \cite{lin2015quantifying}. With the advancement of machine learning (ML), a lot of classical applications (such as drug discovery \cite{yang2022automated}, shallow detection \cite{liao2021shadow}, medical imaging \cite{yang2023device}) have used ML to improve performance, and some techniques are used to combine with ML to enhance the ML effectiveness (such as quantum computing \cite{li2023novel, hu2023battle}). 
In this ML era, data-driven approaches for FWI have emerged as powerful tools\cite{wu2019inversionnet}. ML, leveraging vast datasets, possesses the capability to swiftly and effectively generate velocity maps, enhancing the efficiency of subsurface imaging. However, the application of ML comes with inherent challenges. Unlike physics-driven, which can be applied universally across diverse locations and states, ML exhibits poor performance on unprivileged data \cite{sheng2022larger,sheng2023muffin}.
The crux of the matter lies in the fact that, due to the diversity and complexity of subsurface structures in various locations or the dynamic changes in underground conditions (e.g., CO2 or petroleum leaks), unprivileged data is commonplace in geoscience. 
However, most previous data-driven approaches design the ML model without considering this issue.
As a result, pre-trained ML models often prove ineffective in geoscience applications, necessitating a process of localization.

To compound the challenges, the locations FWI seeking to monitor often present harsh environmental conditions, characterized by limited access to power and network resources \cite{sousa2021geohazards, wang2021lightweight}. As a result, ML models are frequently required to be deployed on edge devices with the constraints of resource limitations and real-time requirements \cite{jiang2020hardware, wu2020intermittent},
necessitate the design of compact models, often achieved through pruning \cite{song2021dancing, wang2023edge}.
This imperative to prune models
comes at a cost. The reduced size 
of pruned models, tailored for real-time applications, result in a dramatic drop in performance, particularly when confronted with unprivileged data. The inherent limitations of edge devices, coupled with the need for real-time processing, pose a formidable challenge to utilize ML models in geoscience applications.

With the recent advent of diffusion models, generative models have acquired enhanced capabilities in generating novel datasets. Initially, it might seem straightforward to leverage generative models, such as Diffusion, for data generation and subsequently fine-tune models to achieve localization. However, the reality is more nuanced. Generative models, like Diffusion, are trained on privileged data, making it challenging for them to effectively generate unprivileged data. Given the limitations of generative models in this regard, a crucial question arises: How do we bridge the gap and generate unprivileged data to address the localization challenges posed by unprivileged data and pruned ML models?

To address these challenges, we propose a novel toolkit, namely EdGeo. Through a fundamental analysis of the challenges of generation and characteristics of the velocity map, we utilize the conditioned generative AI and physics guidance to generate the velocity maps. We then apply the forward model to generate the paired seismic data, enabling the supervised fine-tuning of the ML models.

The main contributions of this paper are outlined as follows.

\vspace{-\topsep}
\setlength{\parskip}{5pt}
\setlength{\itemsep}{0pt plus 1pt}
\begin{itemize}
    \item We introduce the EdGeo toolkit, which incorporates physic-guided optimization. This innovative approach facilitates the generation of high-quality data, particularly in scenarios where data is unprivileged or underrepresented.
    \item {Our approach is tailored to real-world application, emphasizing the need for real-time and adherence to resource constraints. This enables effective localization of the ML model.}
    \item We propose a comprehensive end-to-end fine-tuning framework. It is specifically designed to overcome the localization of the pruned ML model, ensuring its effectiveness and efficiency even in resource-limited environments. 
\end{itemize} 
\vspace{-\topsep}

\begin{figure}[t!]
\begin{center}
\includegraphics[width=\columnwidth]{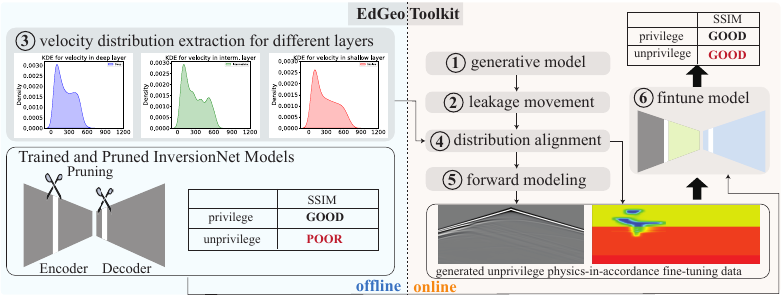}
\caption{Overview of the end-to-end fine-tuning framework and EdGeo toolkit}
\label{fig:framework}
\end{center}
\end{figure}

\vspace{5pt}
\section{Framework}
\label{sec:frame}

As shown in Figure \ref{fig:framework}, the proposed EdeGo has 2 stages: offline and online. 
The offline part utilizes the seismic data and corresponding velocity map to pre-train an InversionNet model.
The velocity distribution at different layers will be obtained according to the unprivileged data or experience. 
The online part is our EdGeo Toolkit, which comprises 6 modules and approaches.

\noindent\textbf{A. Generation Model}

We designed a VM-conditional diffusion to generate a new velocity map based on a given one. 
In the generation process, a pure Gaussian noise will be generated randomly at first. This noise will be denoised gradually and becomes a velocity map at the last step. At each denoising step, the encoded condition will be injected into the U-Net to guide the generation process. The predicted noise will be subtracted from the input, and the changed input will be fed into the U-Net to predict the noise at the next step.

\noindent\textbf{B. Leakage movement}

\begin{figure}[t]
\begin{center}
\includegraphics[width=\columnwidth]{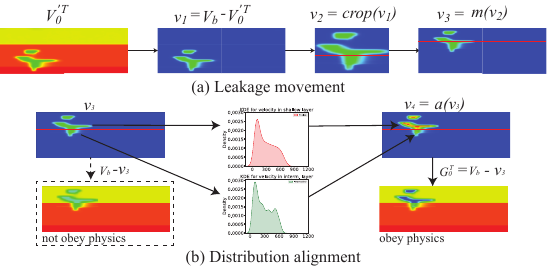}
\caption{Leakage movement and Distribution alignment}
\label{fig:align}
\end{center}
\end{figure}
\vspace{-4pt}

After we obtain the generated velocity map, we aim to move the leakage to the shallow.
Figure \ref{fig:align}(a) shows the details of moving the leakage to the shallow layer.
We first remove the baseline velocity map (velocity map when no leakage) and obtain the pure leakage $v_1 = V_b - V^{'T}_{0}$, where $V_b$ is the baseline velocity map. Then we design a crop function to crop the main leakage area using a threshold $th_l$. A horizontal line will be randomly generated to split the leakage into 2 parts. At last, the randomly generated line will be moved to align the dividing line between the shallow and intermediate layers.

\noindent\textbf{C. Velocity distribution extraction and alignment}

To achieve the localization, we wish to generate the unprivileged data. However, if there is little unprivileged data, the generation model can also not generate the unprivileged data well. After leakage movement, the ideal solution is to align the original distribution at the intermediate or deep layer to the distribution at the shallow layer. 
To achieve this, we opt for a Cumulative Distribution Function (CDF).
Given the velocity variables at shallow $SH$, and intermediate $M$, their CDF functions can be expressed as:
\begin{equation}
\vspace{-5pt}
     F_{SH}(sh) = P(SH \le sh), F_M(m) = P(M \le m)
\label{eq:cdf}
\end{equation}
We hope to find a mapping function from $M$ to $SH$ as $SH = g(M)$.
Once we get the function $g$, we can align $M$ to $SH$.
But the function $g$ is unknown, we need to estimate it according to the known data: $g(m) = sh_1 + \frac{(m - m_1) * (sh_2 - sh_1)}{m_2 - m_1}$,
where $m$ is the velocity we wish to map from $M$ to $SH$, $m1$ and $m2$ are the observed value in $M$, and  $sh_1$ and $sh_2$ are the corresponding values of $m_1$ and $m_2$ in $SH$, and $m_1 \le m \le m_2, sh_1 \le sh_2$. 

However, direct alignment brings the problem of a larger leakage area in the shallow and a smaller leakage area in the intermediate layer. This may lead to the leakage shape shrinking or expanding. 
To solve this, we devise 2 parameters $th_m$ and $th_s$ to filter the no leakage area, and thus Equation \ref{eq:cdf} changes to:
\begin{equation}
\small
\vspace{-5pt}
    F_{SH}(sh) = P(th_s < SH \le sh),
     F_M(m) = P(th_m < M \le m)
\label{eq:cdf2}
\end{equation}

Figure \ref{fig:align}(b) shows the process of distribution alignment. The $v_3$ obtained at the leakage movement step is split into a shallow part and an intermediate part. These 2 parts will be aligned with the shallow distribution and intermediate distribution respectively. At last, it will be recovered by adding the baseline velocity map.

\noindent\textbf{D. Forward Model}

We employ the physics forward modeling:
\begin{equation}
\small
    \nabla^{2} p(r, T) - \frac{1}{V(r)^2}\frac{\partial^{2}p(r, T)}{\partial t^2} = s(r,T)  
    \label{eq:acoustic}
\end{equation}
to produce the paired seismic data.
Based on the second-order central finite difference in the time domain and the Laplacian of the wavefields estimation on the space domain, 
the wave equation can be shown as:
\begin{equation}
\small
\begin{aligned}
    p_{r}^{t+1}=\left(2-v^{2} \nabla^{2}\right) p_{r}^{t}-p_{r}^{t-1}-v^{2}(\Delta t)^{2} s_{r}^{t}
\end{aligned}
    \label{eq:wave}
\end{equation}
Based on this, we can get the seismic data corresponding to the generated velocity map, and use it as the fine-tuning input.

\noindent\textbf{E. Fine-tune model}

For different locations, the occurrence of unprivileged data should be varied. Thus, we propose a loss function $L_f$ to fine-tune the model.
\begin{equation}
\begin{aligned}
\small
    L_f & = \lambda \times (L1(y_u, \hat{y}_u) + L2(y_u, \hat{y}_u)) + \\
    &(1-\lambda) \times (L1(y_p, \hat{y}_p) + L2(y_p, \hat{y}_p)) 
\end{aligned}
    \label{eq:loss}
\end{equation}
where $\hat{y}_u$, $y_u$, $\hat{y}_p$, and  $y_p$ refer to the prediction and ground truth from the unprivileged group and privileged group. $\lambda$ is used to control the effectiveness of generated data.

\section{Experiment}
\label{sec:exp}

\vspace{-1pt}
\noindent\textbf{A. Experimental Setup}

We employ the \textbf{Kimberlina-CO2} \cite{zhou2019data}, a CO$_2$ leakage dataset from 
openFWI \cite{deng2021openfwi}.
We split the data into 2 sets, DM (deep and intermediate) and shallow, according to the leakage area of the group of data. Three metrics (SSIM, mean absolute error (MAE), and mean squared error (MSE)) are used to measure the performance. 
And Raspberry Pi 4 Model B is used as the edge device. 
We employ 3 generative AI competitors for comparison: (1) VAE \cite{kingma2013auto, yang2022making},
(2) VAE-Reg \cite{yang2022making}, and 
(3) Diffusion \cite{ho2020denoising, song2020denoising}.

The InversionNet will be trained using the DM set for 200 epochs. The pruned InvertionNet will be fine-tuned for 120 epochs.
40 epochs are set for the localization fine-tuning. 
And the thresholds $th_s$ and $th_m$ are both set to 50. The $th_l$ is set to the $\frac{1}{3}$ of the maximum value of the velocity map.
We follow  \cite{liu2017learning} to implement pruning.

\noindent\textbf{B. Experimental Results}
\begin{figure}[t]
\begin{center}
\includegraphics[width=\columnwidth]{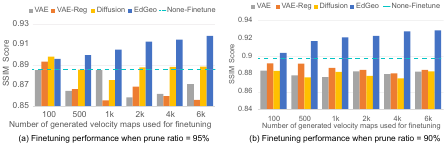}
\caption{Performance with 95\% and 90\% pruning ratio}
\label{fig:bar}
\end{center}
\end{figure}

\noindent\textbf{(1) EdGeo beats the competitors.}
Figure \ref{fig:bar} shows the results of the pruned InversionNet performance fine-tuned by 100, 500, 1000, 2000, 4000, and 6000 generated paired velocity maps and seismic data. In the figure, the grey, red, yellow, and blue bars represent the VAE, VAE-Reg, classical diffusion, and our EdGeo respectively. And the cyan-blue horizontal dash line refers to the SSIM score achieved by the InversionNet before de-biased fine-tuning.
Specifically, Figure \ref{fig:bar} (a) shows the result of a 95\% pruning ratio. We can observe that as the number of used EdGeo generation data grows, the SSIM score of InversionNet increases. At the number of 100 generated data used, the InversionNet fine-tuned by EdGeo achieved 0.8963, which is only 0.0003 lower than Diffusion, but higher than the VAE and VAE-Reg. Except for this group, the InversionNet fine-tuned by EdGeo got the highest SSIM score compared with the competitors. For the group of 6000 pair data, InversionNet fine-tuned by EdGeo gained a 0.9188 SSIM score, which is 3.37\% higher than the Diffusion and 7.31\% higher than the VAE-Reg. The SSIM scores obtained by EdGeo fine-tuning are all higher the the none-finetune. However, the data generated by the competitors may damage the performance of InversionNet, observing a lot of bars lower than the none-finetune line.
Similar results can be found when the pruning ratio equals 90\%. At a 90\% pruning ratio, the InversionNet fine-tuned by EdGeo achieves the highest SSIM score at all groups with different numbers of used data. When 4000 pairs of generated data are used, the InversionNet fine-tuned by EdGeo gained a 0.9276 SSIM score, 5.99\% higher than the one fine-tuned by Diffusion. When 6000 data are used, EdGeo fine-tuning got a 0.9289 SSIM score, which is 3.60\% higher than the none-finetune.

\begin{figure}[t]
\begin{center}
\includegraphics[width=\columnwidth]{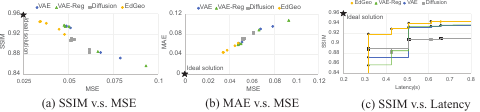}
\caption{Comparison between the EdGeo and competitors}
\label{fig:metrics}
\end{center}
\end{figure}

\noindent\textbf{(2) EdGeo achieves best in different pruning ratios.}
SSIM concentrates on the structure while ignoring the details. However, the FWI tasks need to pay attention to pixel-level accuracy. Thus, we also bring the MAE and MSE to evaluate the fine-tuned InversionNet performance. 
In the second set of experiments, we demonstrate that EdGeo achieves the best performance
among SSIM, MSE, and MAE, MSE, compared with the competitors, with different pruning ratios varying from 75\% to 95\%. 
Figure \ref{fig:metrics} reports the 3 kinds of metrics results.
In Figure \ref{fig:metrics}(a),
the x-axis is the MSE and the y-axis is the SSIM score.
The ideal solution is located in the left-up corner, shown as a star. 
The yellow circle points correspond to EdGeo fine-tuning, 
while other color and shape points refer to kinds of competitors. 
This figure clearly shows that EdGeo achieves the highest SSIM with the lowest MSE. All EdGeo points are concentrated at the left-up part, compared with competitors.
And we can observe that, compared with SSIM, EdGeo gains more benefits on MSE. 
When it comes to Figure \ref{fig:metrics} (b), the x-axis is the MSE and the y-axis is the MAE. The ideal solution changes to the left-down corner.  Figure \ref{fig:metrics}(b) consistently shows that EdGeo achieves the lowest MSE and MAE at the same time. 

Although VAE-Reg improves the SSIM score when the pruning ratio is at 85\%, it may enhance the prediction ability of the leakage at the intermediate and deep layers, and may not predict the leakage at the shallow layer. The MAE and MSE results are evidence, and the sub-section \ref{subsec:vis} reports the visualization result.

\noindent\textbf{(3) Performance and Latency}
Figure \ref{fig:metrics} (c) shows the inference latency of the pruned InversionNet with different ratios. 
The x-axis represents the latency (in seconds) on Raspberry Pi and the y-axis is the SSIM score. The ideal solution is located in the left-up corner, meaning a higher performance and a lower latency. EdGeo significantly pushes forward the Pareto frontiers in the trade-off SSIM  and latency. Considering the real-time requirement, only the 90\% and 95\% pruned models can be accepted if the latency requirement is below 0.5s, which also proves the value of EdGeo.

\noindent\textbf{C. Result Visualization}
\vspace{-1pt}
\label{subsec:vis}

\begin{figure}[t]
\begin{center}
\includegraphics[width=\columnwidth]{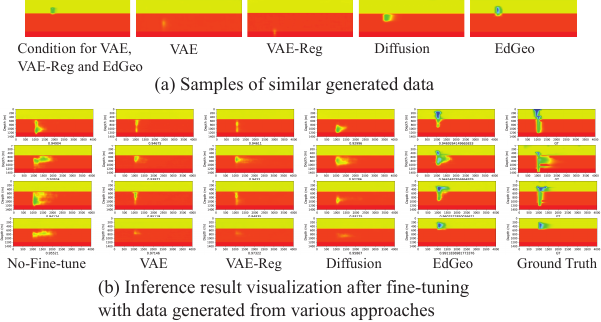}
\caption{Samples of similar generated data, and inference result visualization after fine-tuning}
\label{fig:gen}
\end{center}
\vspace{-4pt}
\end{figure}

Figure \ref{fig:gen} (a) shows similar generation velocity maps from different approaches. Specifically, the first velocity map shown in Figure \ref{fig:gen} (a) is used as a part of the condition for VAE and VAE-Reg (VAE and VAE-Reg require 2 continuous-time leakages as conditions), and the condition of EdGeo. As the figures show, VAE, VAE-Reg, and Diffusion can not generate the leakage at the shallow layer. As well, the shape of velocity map generated by EdGeo is similar to the condition, but with greater leakage. This is also the reason why we chose the conditional diffusion model because we wish to generate the leakage with a similar shape as the condition velocity map. This is a key to the localization.

Figure \ref{fig:gen} (b) shows the inference result from all approaches when the ratio is 85\%. In Figure \ref{fig:gen} (b), the velocity maps in the first column refer to the inference result without fine-tuning using the generated data. The velocity maps in the second to fifth columns are inference results from InversionNet fine-tuned by VAE, VAE-Reg, Diffusion, and EdGeo. The last column reports the ground truth. We can observe that EdGeo can predict the leakage at the shallow layer, while other approaches can not. Although VAE-Reg improves performance, it actually enhances the prediction of intermediate and deep layers.
However, this falls short of our ultimate goal. What we truly aim for is localization—specifically, the ability to predict the leakage at the shallow layer. And EdGeo achieves this.

\section{Conclusion}

\label{sec:con}
In this paper, we propose the EdGeo aiming to address the common challenge of unprivileged data existing in the geoscience FWI tasks. Given the CO$_2$ leakage monitoring task, EdGeo first utilizes the conditional diffusion model to generate the velocity map which may be similar to the condition. According to the leakage movement and distribution alignment, we can generate the unprivileged data, i.e., leakage at the shallow layer. The physics forward modeling produces corresponding seismic data, enabling the fine-tuning of pruned ML models. The experimental result shows that the ML model fine-tuned by the data generated by the EdGeo has the ability to predict the leakage at the shallow layer, and achieves better metrics performance compared with competitors.

{\tiny
\bibliography{paper.bib}

\begin{thebibliography}{10}
\providecommand{\url}[1]{#1}
\csname url@samestyle\endcsname
\providecommand{\newblock}{\relax}
\providecommand{\bibinfo}[2]{#2}
\providecommand{\BIBentrySTDinterwordspacing}{\spaceskip=0pt\relax}
\providecommand{\BIBentryALTinterwordstretchfactor}{4}
\providecommand{\BIBentryALTinterwordspacing}{\spaceskip=\fontdimen2\font plus
\BIBentryALTinterwordstretchfactor\fontdimen3\font minus \fontdimen4\font\relax}
\providecommand{\BIBforeignlanguage}[2]{{%
\expandafter\ifx\csname l@#1\endcsname\relax
\typeout{** WARNING: IEEEtran.bst: No hyphenation pattern has been}%
\typeout{** loaded for the language `#1'. Using the pattern for}%
\typeout{** the default language instead.}%
\else
\language=\csname l@#1\endcsname
\fi
#2}}
\providecommand{\BIBdecl}{\relax}
\BIBdecl

\bibitem{tran2013sinkhole}
K.~T. Tran and et~al., ``Sinkhole detection using 2d full seismic waveform tomography,'' \emph{Geophysics}, 2013.

\bibitem{alumbaugh2023kimberlina}
D.~Alumbaugh and et~al., ``The kimberlina synthetic multiphysics dataset for co2 monitoring investigations,'' \emph{Geoscience Data Journal}, 2023.

\bibitem{lin2015quantifying}
Y.~Lin and et~al., ``Quantifying subsurface geophysical properties changes using double-difference seismic-waveform inversion with a modified total-variation regularization scheme,'' \emph{Geophysical Supplements to the Monthly Notices of the Royal Astronomical Society}, 2015.

\bibitem{yang2022automated}
J.~Yang and et~al., ``Automated architecture search for brain-inspired hyperdimensional computing,'' \emph{arXiv preprint arXiv:2202.05827}, 2022.

\bibitem{liao2021shadow}
J.~Liao and et~al., ``Shadow detection via predicting the confidence maps of shadow detection methods,'' in \emph{Proceedings of the 29th ACM International Conference on Multimedia}, 2021, pp. 704--712.

\bibitem{yang2023device}
J.~Yang and et~al., ``On-device unsupervised image segmentation,'' \emph{arXiv preprint arXiv:2303.12753}, 2023.

\bibitem{li2023novel}
J.~Li and et~al., ``A novel spatial-temporal variational quantum circuit to enable deep learning on nisq devices,'' in \emph{2023 IEEE International Conference on Quantum Computing and Engineering (QCE)}, vol.~1.\hskip 1em plus 0.5em minus 0.4em\relax IEEE, 2023, pp. 272--282.

\bibitem{hu2023battle}
Z.~Hu and et~al., ``Battle against fluctuating quantum noise: Compression-aided framework to enable robust quantum neural network,'' \emph{arXiv preprint arXiv:2304.04666}, 2023.

\bibitem{wu2019inversionnet}
Y.~Wu and et~al., ``Inversionnet: An efficient and accurate data-driven full waveform inversion,'' \emph{IEEE Transactions on Computational Imaging}, 2019.

\bibitem{sheng2022larger}
Y.~Sheng and et~al., ``The larger the fairer? small neural networks can achieve fairness for edge devices,'' in \emph{Proceedings of the 59th ACM/IEEE Design Automation Conference}, 2022, pp. 163--168.

\bibitem{sheng2023muffin}
Y.~{}Sheng and et~al., ``Muffin: A framework toward multi-dimension ai fairness by uniting off-the-shelf models,'' in \emph{2023 60th ACM/IEEE Design Automation Conference (DAC)}.\hskip 1em plus 0.5em minus 0.4em\relax IEEE, 2023, pp. 1--6.

\bibitem{sousa2021geohazards}
J.~J. Sousa and et~al., ``Geohazards monitoring and assessment using multi-source earth observation techniques,'' \emph{Remote Sensing}, 2021.

\bibitem{wang2021lightweight}
Z.~Wang, Y.~Wu, and et~al, ``Lightweight run-time working memory compression for deployment of deep neural networks on resource-constrained mcus,'' in \emph{Proceedings of the 26th Asia and South Pacific Design Automation Conference}, 2021, pp. 607--614.

\bibitem{jiang2020hardware}
W.~Jiang and et~al., ``Hardware/software co-exploration of neural architectures,'' \emph{IEEE Transactions on Computer-Aided Design of Integrated Circuits and Systems}, vol.~39, no.~12, pp. 4805--4815, 2020.

\bibitem{wu2020intermittent}
Y.~Wu and et~al, ``Intermittent inference with nonuniformly compressed multi-exit neural network for energy harvesting powered devices,'' in \emph{2020 57th ACM/IEEE Design Automation Conference (DAC)}.\hskip 1em plus 0.5em minus 0.4em\relax IEEE, 2020, pp. 1--6.

\bibitem{song2021dancing}
Y.~Song and et~al., ``Dancing along battery: Enabling transformer with run-time reconfigurability on mobile devices,'' in \emph{2021 58th ACM/IEEE Design Automation Conference (DAC)}.\hskip 1em plus 0.5em minus 0.4em\relax IEEE, 2021, pp. 1003--1008.

\bibitem{wang2023edge}
Z.~Wang, I.~Putla, W.~Jiang, and Y.~Lin, ``Edge-inversionnet: Enabling efficient inference of inversionnet on edge devices,'' in \emph{Third International Meeting for Applied Geoscience \& Energy}.\hskip 1em plus 0.5em minus 0.4em\relax Society of Exploration Geophysicists and American Association of Petroleum~…, 2023, pp. 1059--1063.

\bibitem{zhou2019data}
Z.~Zhou and et~al., ``A data-driven co2 leakage detection using seismic data and spatial--temporal densely connected convolutional neural networks,'' \emph{International Journal of Greenhouse Gas Control}, 2019.

\bibitem{deng2021openfwi}
C.~Deng and et~al., ``{OpenFWI:} large-scale multi-structural benchmark datasets for seismic full waveform inversion,'' \emph{arXiv preprint arXiv:2111.02926}, 2021.

\bibitem{kingma2013auto}
D.~P. Kingma and et~al., ``Auto-encoding variational bayes,'' \emph{arXiv preprint arXiv:1312.6114}, 2013.

\bibitem{yang2022making}
Y.~Yang and et~al., ``Making invisible visible: Data-driven seismic inversion with spatio-temporally constrained data augmentation,'' \emph{IEEE Transactions on Geoscience and Remote Sensing}, 2022.

\bibitem{ho2020denoising}
J.~Ho and et~al., ``Denoising diffusion probabilistic models,'' \emph{Advances in neural information processing systems}, 2020.

\bibitem{song2020denoising}
J.~Song and et~al., ``Denoising diffusion implicit models,'' \emph{arXiv preprint arXiv:2010.02502}, 2020.

\bibitem{liu2017learning}
Z.~Liu and et~al., ``Learning efficient convolutional networks through network slimming,'' in \emph{Proceedings of ICCV}, 2017.

\end{thebibliography}
\bibliographystyle{IEEEtran}}

\end{document}